\documentclass[a4paper]{article}

\usepackage[margin=1in]{geometry}
\usepackage{microtype}
\usepackage{graphicx}
\usepackage{subfigure}
\usepackage{booktabs} %
\usepackage{natbib}

\usepackage[dvipsnames]{xcolor}         %
\usepackage[colorlinks=true,linkcolor=violet,urlcolor=black,citecolor=MidnightBlue,allbordercolors={1 1 1}]{hyperref}       %

\usepackage{amsmath}
\usepackage{amssymb}
\usepackage{mathtools}
\usepackage{amsthm}
\usepackage{enumitem}
\usepackage{listings}
\usepackage[capitalize,noabbrev]{cleveref}
\usepackage{xspace}

\def\singlecolumn{1}

\input{def.set}
\input{def-local.set}
\renewcommand{\vshrink}[1]{}

\usepackage[textsize=tiny]{todonotes}
\title{\titlename}
\author{Ziyu Wang\\
\normalsize University of Oxford \\
\normalsize \email{wzy196@gmail.com} 
\and 
Chris Holmes \\ 
\normalsize University of Oxford \\
\normalsize \email{cholmes@stats.ox.ac.uk}
}
\date{}

\begin{document}

\maketitle

\begin{abstract}
    Applications of large language models often involve the generation of free-form %
responses, in which case uncertainty quantification becomes challenging. 
This is 
due to the need to identify task-specific uncertainties (e.g., about the semantics) which appears difficult to define in general cases.
This work addresses these challenges from a perspective of Bayesian decision theory, starting from the assumption that 
our utility is characterized by a similarity measure that compares a generated response with a hypothetical true response.
We discuss how 
this assumption enables principled quantification of the model's subjective uncertainty and its calibration. %
We further derive a measure for epistemic uncertainty, based on 
a missing data perspective and its characterization as an excess risk. 
The proposed methods can be applied to black-box language models.  
We illustrate the methods on question answering and machine translation tasks. 
Our experiments provide a principled evaluation of task-specific calibration, and demonstrate that epistemic uncertainty offers a promising deferral strategy for efficient data acquisition in %
in-context learning.

\end{abstract}

\section{Introduction}\label{sec:intro}

We are interested in uncertainty quantification (UQ) for language models (LMs) in \emph{free-form} natural language generation (NLG): %
given instruction $I$, the model generates a response $y'$ based on its predictive distribution $p_M(y\mid I)$, where $y'$ could be \emph{any} natural language passage that fits the instruction. 
An example of this is in question answering (QA): 
given a %
question from the user, 
the model may provide a brief answer, but 
it may also follow with supporting facts and explanations, which can vary in form and detail. %
The user can be satisfied by a wide variety of responses, irrespective of their style or (to some extent) the choice of supporting facts included. 

Free-form NLG poses significant challenges to uncertainty quantification: 
some aspects of generation are irrelevant to the task's purpose and best excluded from uncertainty quantification, but it often appears that 
we are unable to characterize them 
precisely. 
If left unaddressed, however, the model's variation %
in the irrelevant aspects may dominate in standard uncertainty measures such as token-level entropy \citep{kuhn_semantic_2023}, making them uninformative about %
the model's actual performance on the task. 

Starting from \citet{kuhn_semantic_2023}, 
a recent line of work \citep{kuhn_semantic_2023,lin2024generating,zhang_sac3_2023,aichberger_how_2024} 
studied this issue and %
proposed measuring the ``semantic uncertainty'' of generation; ``semantics'' is defined as the equivalence class of textual responses that logically entail one another. %
Empirical improvements in downstream tasks evidenced their contributions and 
highlighted the importance of \emph{task-specific} uncertainty quantification, 
but important conceptual and practical issues remain. 
From a practical perspective, semantic equivalence is estimated using machine learning models, resulting in imprecise estimates that do not necessarily define an equivalence relation. The imprecision necessitates the introduction of heuristics to post-process the estimates or to aggregate them through other means. 
Conceptually, the notion of semantic equivalence does not always provide a valid or complete characterization of relevance for certain tasks. 
Style transfer \citep{jin2022deep} tasks provide an example of the first kind.  
For the second scenario, consider a QA task where 
the model is fully certain about the answer to a question, 
as well as a large set of supporting facts, each of which provides complete and independent justification for the answer; yet there is still the ``uncertainty'', or variation across generations, about which facts are included in a response. Then 
very few of the model generations 
may logically entail one another, and there is a very high level of ``semantic uncertainty'' just 
as if there was complete uncertainty 
about the correct answer. 

This paper is about the observation that the above challenges %
can be resolved in a more general setup from a perspective of \emph{Bayesian decision theory} \citep{savage1972foundations}. 
We start by 
assuming that we can compute a similarity-type %
measure $S(y', y; I)$ that characterizes our \emph{utility function} when the model generates $y'$ in respond to $I$ and $y$ is a (true or hypothetical) correct response.
This generalizes previous works based on semantic equivalence, which corresponds to defining $S$ using entailment. 
It can also be applied to traditional structured prediction tasks where a choice of $S$ is readily available (e.g., %
lexical scores for machine translation, \citealp{koehn2009statistical}), as well as any 
task where evaluation can be %
implemented using LMs that compute $S$ based on few-shot demonstrations or detailed instructions \citep[e.g.,][]{yu2023evaluating,wang2023chatgpt,Bannur2024MAIRA2GR}.

To quantify the subjective uncertainty defined by the LM, 
we 
assume the generation $y$ is always chosen to %
maximize the expected utility under the LM's predictive distribution $p_M$ (\S\ref{sec:setup}). 
This is a common, albeit implicit, assumption in NLG \citep{bickel2015mathematical,bertsch_its_2023}.  
It can be supported by a belief that high-capacity LMs may approximate Bayesian inference \citep[e.g.,][]{xie2021explanation,akyurek2022learning,hahn2023theory}, 
and is also more broadly applicable whenever we view $p_M$ as our best available model for $y\mid I$. 
Subjective uncertainty is then naturally characterized by the Bayes risk, %
or equivalently the maximum achievable expected utility, where the action space is defined by the candidate generations available.  
This simple observation allows us to understand that previous methods for ``semantic uncertainty'' can be adapted to a broader range of scenarios, regardless of whether semantic equivalence is relevant. It also highlights a unique, principled approach for aggregating similarity measures among generations (\S\ref{sec:uq-start}). %

For subjective uncertainty measures to be useful, it will be ideal if the LM is more calibrated. 
It has been unclear in previous works how calibration can be evaluated in general %
NLG tasks, given the distinction between relevant and irrelevant differences for a %
task and our apparent inability to formally characterise it. %
The decision-theoretic view 
provides a natural answer to this question: 
an LM is deemed to possess a calibrated notion of uncertainty if the expected subjective utility of its action matches the expectation of the actually incurred utility %
w.r.t.~the true data distribution (\S\ref{sec:calibration}). 
This observation allows us to quantify the calibration of LMs through reliability diagrams \citep{murphy1977reliability} and a generalization of the expected calibration error \citep{naeini2015obtaining}. 

Bayesian modelling enables the decomposition of %
predictive uncertainty into 
\emph{epistemic uncertainty} and \emph{aleatoric uncertainty} \citep{der2009aleatory}. 
Epistemic uncertainty refers to the reducible proportion of uncertainty and can guide data acquisition \citep{kendall2017uncertainties}, or more broadly, indicate when additional information
can be most effectively used to improve prediction. 
The quantification of epistemic uncertainty is of interest in many LM applications, as predictions can often benefit from additional information \citep{lewis2020retrieval,brown2020language,li_eliciting_2023}, which may be costly to obtain \citep{agarwal_many-shot_2024,min2024silo}. 
Yet it appears challenging due to the black-box nature of LMs and the free-form generations involved. 
As we show in \S\ref{sec:eu}, the decision-theoretic view leads to a principled measure of task-specific epistemic uncertainty in in-context learning (ICL, \citealp{brown2020language}); 
key additional ingredients include a missing data perspective to Bayesian modelling \citep{fong_martingale_2021} 
and a connection between epistemic uncertainty and an excess risk \citep{xu_minimum_2020}. 
The resulted measure is connected to and generalizes several recent methods, %
which provides independent justification for its use. 

The methodological developments allow us to better understand the ``probabilistic uncertainty'' in modern LMs, that is, the uncertainty reflected in the variation of the model's predictive distribution. 
\S\ref{sec:exp} provides experimental illustrations: 
We revisit past comparisons \citep{tian_just_2023,lin2024generating} of probabilistic and non-probabilistic UQ %
for instruction-tuned LMs by conducting 
a principled evaluation of task-specific calibration (\S\ref{sec:exp-qa}). 
In \S\ref{sec:exp-mt}, we demonstrate that epistemic uncertainty provides a deferral strategy \citep{madras2018predict,dohan2022language} that facilitates efficient data acquisition in many-shot ICL \citep{agarwal_many-shot_2024}. %

The rest of this paper is structured as follows: \S\ref{sec:method} presents the proposed methodology, \S\ref{sec:related-work} reviews related work, \S\ref{sec:exp} presents the experiments, and \S\ref{sec:conclusion} provides concluding remarks.

\section{Quantifying Uncertainty and Calibration}\label{sec:method}

\subsection{A Utilitarian Setup}\label{sec:setup}

The setup of this paper is as follows: 
\ifdefined\doublecolumn
\begin{itemize}[leftmargin=*,topsep=2pt,itemsep=2pt]
\else
\begin{itemize}
\fi
    \item 
we have a NLG task: given prompt $I\in\cI$, generate a response $y'\in\cY$ where $\cY$ denotes the space of natural language responses; 
\item
our utility can be quantified through a similarity measure $S(y', y; I) \in \RR$, which can be cheaply evaluated;
\item %
we have %
an LM with predictive distribution $p_M$, and, in the absence of further evidence, consider $p_M$ sufficiently trustworthy so that the ideal generation (i.e., \emph{action}) $y'$ for $I$ should maximize the \emph{expected utility}, 
\begin{equation}\label{eq:expected-loss}
y' := \argmax_{y'\in\actionSpace} \EE_{y\sim p_M(\cdot\mid I)} S(y', y; I),
\end{equation}
where $\cY'_I\subset \cY$ is %
the space of candidate generations. %
\end{itemize}

In the above, $p_M$ represents a subjective belief about the true data distribution. 
$\cY_I'\subset \cY$ can be %
determined based on computational constraints, e.g., as a separate set of samples from $p_M(\cdot\mid I)$.\footnote{
When $\cY_I'$ is stochastic, expectation %
quantities such as \eqref{eq:expected-loss} should also be averaged over the randomness in 
$\cY_I'$. For brevity, we will omit all such (outmost) expectations.
}
The utility $S$ determines a \emph{risk} function $r(y',y;I) = -S(y',y; I)$, which can be scaled and shifted as desired.

As discussed in \S\ref{sec:intro}, this setup is very general. 
This decision-theoretic perspective is taken in the framework of minimum Bayes risk (MBR) decoding \citep{bickel2015mathematical}, which advocates for generating $y$ based precisely on \eqref{eq:expected-loss}. MBR recovers many modern methods for natural language generation \citep{bertsch_its_2023}, which %
can be viewed as implicitly adopting %
the same view. %
Note that %
the decision-theoretic setup does not compel us to assume that $p_M$ is strictly a Bayesian model \citep[e.g., in the sense of][]{xie2021explanation}: although that %
will provide a good justification, it is also possible that a non-Bayesian LM better approximates our %
belief than any available Bayesian model does.

\subsection{Task-Specific Measures of Subjective Uncertainty}\label{sec:uq-start}

It is well known that %
Bayesian uncertainty measures are connected to minimum achievable risks of the form 
\begin{equation}\label{eq:mar}
\BayesRisk := \min_{y'\in\actionSpace} \EE_{y\sim p_M(\cdot\mid I)} r(y',y;I).
\end{equation}
For example, if $\cY=\actionSpace[I]\subset\RR$ and $r(y,y';I)=(y-y')^2$, 
$\BayesRisk$ will be equivalent to the predictive variance; %
if with an abuse of notation we redefine $y'$ as a %
density function and 
$r(y',y;I) \gets \log y'(y)$, $R_{B,\cY'_I}(I)$ %
will become the entropy. 
When the action space $\cY'_I\subsetneq \cY$ is a strict subset, 
\eqref{eq:mar} becomes connected to measures of ``usable information'' discussed in \citet{xu2020theory}. %

It is thus natural to also adopt \eqref{eq:mar} as a measure of subjective uncertainty in NLG. 
By definitions, it represents the \emph{task-specific} uncertainty we are interested in, and we recover measures of semantic uncertainty if we define $r$ based on semantic equivalence. 
\eqref{eq:mar} can be implemented through Monte-Carlo estimation.

A common generation strategy is defined by the Gibbs predictor, which corresponds to \eqref{eq:expected-loss} with $\actionSpace := \{y_I\}$, where $y_I\sim p_M(\cdot\mid I)$ is a single random sample. In such cases, the expected uncertainty $\EE_{\actionSpace} \BayesRisk$ is connected to the degree statistic in \citet[Eq.~8]{lin2024generating}; see also \citet[p.8]{wang2022self}. The difference is that we use a similarity measure that we assume to define our utility. Compared with previous works \citep{kuhn_semantic_2023,lin2024generating,nikitin2024kernel} which explored many uncertainty measures based on the aggregation of similarity measures, the decision-theoretic perspective highlights one unique method without any underspecified design choices, justifies its use beyond the scope of semantic equivalence, and shows that different choices of predictors ($\cY_I'$) should be matched with different uncertainty measures. 
Although being a measure of subjective uncertainty, it is not guaranteed to always outperform 
heuristic alternatives on downstream tasks, in particular if the LM is severely miscalibrated. 

\subsection{Evaluation of Task-Specific Calibration}\label{sec:calibration}

Subjective uncertainty measures are most useful when the LM is reasonably calibrated. 
As discussed in \citet{huang_uncertainty_2024}, 
many previous works studying uncertainty in free-form NLG did not distinguish between calibration and predictive performance in their evaluation, 
which leads to a relative lack of understanding about the calibration %
in such settings. 

A common definition of calibration is as follows \citep[see e.g.,][]{johnson_experts_2024}: 
suppose the spaces $\cY,\cI$ are discrete, and denote the true data distribution as $p_0$; a model $p_M$ is considered calibrated w.r.t.~some grouping function $G(I): \cI\to\cG$ if we have
\ifdefined\singlecolumn
\begin{equation}\label{eq:calibration-full}
\EE_{p_0}(p_M(y=y'\mid I) \mid G(I)=g) = 
\EE_{p_0}(p_0(y=y'\mid I) \mid G(I)=g) ~~\forall g\in\cG,y'\in\cY.
\end{equation}
\else
\begin{align}
&\phantom{=}\EE_{p_0}(p_M(y=y'\mid I) \mid G(I)=g) \nonumber \\
&= \EE_{p_0}(p_0(y=y'\mid I) \mid G(I)=g) ~~\forall g\in\cG,y'\in\cY.\label{eq:calibration-full}
\end{align}
\fi
$G(I)$ can be defined based on uncertainty measures, or it may incorporate additional information such as input demographics. 
In NLG, however, %
\eqref{eq:calibration-full} can be trivially violated if, e.g., the LM always generates a more verbose response compared with $p_0$, 
which may happen due to instruction tuning \citep{adlakha2023evaluating}.

In light of the assumptions in \S\ref{sec:setup}, 
it is natural to relax \eqref{eq:calibration-full} to require that 
for all $g\in\cG, y'\in\actionSpace[I], s\in\RR$ 
\ifdefined\singlecolumn
\begin{align}
\EE(p_M(\{y: S(y', y;\cI)=s\}\mid I) \mid G(I)=g) &= 
\EE(p_0(\{y: S(y', y;\cI)=s\}\mid I) \mid G(I)=g)  \label{eq:calibration-NLG}\tag{\ref{eq:calibration-full}'}
\end{align}
\else
\begin{align}
&\phantom{=}\EE(p_M(\{y: S(y', y;\cI)=s\}\mid I) \mid G(I)=g)  \nonumber 
\\ &= 
\EE(p_0(\{y: S(y', y;\cI)=s\}\mid I) \mid G(I)=g)  \label{eq:calibration-NLG}\tag{\ref{eq:calibration-full}'}
\end{align}
\fi
\eqref{eq:calibration-NLG} is equivalent to \eqref{eq:calibration-full} 
if $\cY'_I=\cY$ and $S(y',y;\cI) = \mbf{1}\{y=y'\}$, or more generally if for all $I\in\cI$ the map from $y$ to the function 
$$ %
s_{y,I}: \cY_I'\to\RR,~ y'\mapsto S(y',y;I)
$$
 is injective. %
\eqref{eq:calibration-NLG} becomes weaker if $S$ %
only depends on $y$ and $y'$ through a non-invertible function, e.g., a map from a natural language passage to its \emph{semantics}. 
In such cases, it is clear that \eqref{eq:calibration-NLG} remains sufficiently strong while focusing on task-specific calibration.

\begin{remark}
We can generally view the function $s_{y,I}(y')$ as a representation of the ``task-specific semantics''.\footnote{
For the purposes of MBR generation %
and uncertainty quantification, it suffices to define the function $s_{y,I}$ on $\actionSpace[I]$. 
For other purposes %
it may be helpful to extend its domain. %
} 
In fact, the replacement of %
$y'\in\actionSpace[I]$ with %
$s_{y',I}(\cdot)$ has been advocated by \citet{savage1972foundations}, and $s_{y,I}$ is referred to as a \emph{Savage act} \citep[see e.g.,][\S 2.4]{marinacci2015model}. 
\end{remark}

A main reason we are interested in %
\eqref{eq:calibration-full} or \eqref{eq:calibration-NLG} is 
that they guarantee 
the calibration of uncertainty measures. It follows from \eqref{eq:calibration-NLG}  that 
$
\EE_{p_0(I \mid G(I)=g)p_M(y\mid I)} r(y', y; I) = 
\EE_{p_0(I \mid G(I)=g)p_0(y\mid I)} r(y', y; I)
$ for all $y'$, and thus 
\begin{align*}
\EE_{p_0(I\mid G(I)=g)}\BayesRisk = \EE_{p_0(I\mid G(I)=g)p_0(y\mid I)} r(\hat y_I, y; I),
\end{align*}
where $\hat y_I$ denotes the MBR generation \eqref{eq:expected-loss}. 
Plugging in $G(I) = \BayesRisk$, we find that the calibration criterion %
\eqref{eq:calibration-NLG} holds only if 
\ifdefined\singlecolumn
\begin{equation}\label{eq:gece}
\text{ECE}(p_M) := \EE_{s} |f_M(s) - s| = 0,  %
~~~\text{where }
f_M(s) := \EE_{I,y,\hat y_I}
(r(\hat y_I,y; I) \mid \BayesRisk=s)~\forall s\in\RR.
\end{equation}
\else
\begin{equation}\label{eq:gece}
\begin{aligned}
&\text{ECE}(p_M) := \EE_{s} |f_M(s) - s| = 0,  %
~~~\text{where } \\
&f_M(s) := \EE_{I,y,\hat y_I}
(r(\hat y_I,y; I) \mid \BayesRisk=s)
\end{aligned}
\end{equation}
for all $s\in\RR$.
\fi
\eqref{eq:gece} generalizes the \emph{expected calibration error} (ECE, \citealp{naeini2015obtaining}) in classification, which it recovers with $r(y',y;I) \gets \mbf{1}\{y'\ne y\}$. As in the classification case, we can estimate \eqref{eq:gece} through histogram binning. The plot for $f_M$ is called a \emph{reliability diagram} \citep{murphy1977reliability} and provides insights on overconfidence ($f_M(s)<s$) or underconfidence. Importantly, these methods remain applicable in NLG without the need to introduce any binary notion of correctness. Moreover, they will automatically focus on the task-specific aspects of generation, as $\text{ECE}$ and $f_M$ only depend on model samples through the utility.

\subsection{Representing and Eliciting Epistemic Uncertainty}\label{sec:eu}

We now turn to the quantification of epistemic uncertainty, using ICL as a key example. 
Epistemic uncertainty is the proportion of uncertainty that can be reduced given additional observations. 
It is particularly relevant when such observations are costly to obtain or incorporate into the generation process. 
In these scenarios, we can limit their use to queries with a higher level of \emph{reducible} (i.e., epistemic) uncertainty. 

\vshrink{-0.3em}
\paragraph{Epistemic uncertainty in ICL.}
In ICL the LM is prompted with $I=\<z_{1:n},x_*\>=\<x_1,y_1,\ldots,x_n,y_n,x_*\>$, where 
$z_i:=(x_i,y_i)$ are demonstration input-output pairs and $x_*$ denotes the 
test query. %
$n$ can be arbitrary.  
We assume that the user's utility does not depend on $z_{1:n}$ given $x_*$, so that the risk has the form $r(y,y';I) \equiv r(y,y'; x_*)$. 
We propose 
the following measure of epistemic uncertainty, inspired by \citet{fong_martingale_2021,xu_minimum_2020}:
\ifdefined\singlecolumn
\begin{align*}
 & 
\inf_{y'\in\actionSpace[I]}
    \EE_{p_x(x_{n+1:N}\mid I)}\EE_{p_M(y_{n+1:N}\mid y_{1:n}, x_{1:N})}
         \EE_{p_M(y_*\mid z_{1:N},x_*)} r(y',y_*; x_*)
          \\
&\hspace{4em}
- \EE_{p_x(x_{n+1:N}\mid I)}\EE_{p_M(y_{n+1:N}\mid y_{1:n}, x_{1:N})}
        \inf_{y'\in\actionSpace[I]} \EE_{p_M(y_*\mid z_{1:N},x_*)} r(y',y_*; x_*).
\numberthis\label{eq:eu}
\end{align*}
\else
\begin{align*}
&\inf_{y'\in\actionSpace[I]}
    \EE_{p_x(x_{n+1:N}\mid I)}\EE_{p_M(y_{n+1:N}\mid y_{1:n}, x_{1:N})} \\ 
&\hspace{4em}
         \EE_{p_M(y_*\mid z_{1:N},x_*)} r(y',y_*; x_*) \\
&
- \EE_{p_x(x_{n+1:N}\mid I)}\EE_{p_M(y_{n+1:N}\mid y_{1:n}, x_{1:N})}\\
&\hspace{4em}
        \inf_{y'\in\actionSpace[I]} \EE_{p_M(y_*\mid z_{1:N},x_*)} r(y',y_*; x_*).
\numberthis\label{eq:eu}
\end{align*}
\fi
In the above, $p_x$ denotes a user-specified distribution of inputs, 
$y_{n+1:N}\sim \prod_{j=n+1}^N p_M(y_j\mid z_{1:j-1},x_j)$ follow the LM's %
predictive distribution, %
and $N>n$ is a hyperparameter. %
\eqref{eq:eu} can be estimated through autoregressive sampling and Monte-Carlo estimation; %
see \S\ref{sec:exp-mt} for implementation details. 

Comparing with \eqref{eq:mar}, we can see that \eqref{eq:eu} is the difference between the total uncertainty from two predictors; 
the second predictor has access to the additional demonstrations 
$z_{n+1:N}=(x_{n+1:N}, y_{n+1:N})$, where %
$y_{n+1:N}$ are model generations. 
Thus, 
\eqref{eq:eu} is a subjective measure of the %
\emph{reducible uncertainty} given $x_{n+1:N}$. %
By convexity of the $\inf$ functional, \eqref{eq:eu} is always non-negative; so is its Monte-Carlo estimate as long as we use the same set of samples for %
its both terms. 

\paragraph{Bayesian justifications.} 
To further understand \eqref{eq:eu}, 
consider a simplified setting as follows: suppose $p_M$ is equivalent to a Bayesian model that assumes $z_{1:n}$ are i.i.d.~conditional on a latent variable $\theta$, $x_{n+1:N}\sim p_x(\cdot\mid I)$ are conditionally i.i.d., 
and %
$z_{n+1:N}$ are sufficiently informative %
so that the likelihood function $p(y=\cdot\mid x=x_*,\theta)$ determined by $\theta\mid z_{1:N}$ is identifiable given infinite samples. Then under mild technical conditions, as $N\to\infty$ \eqref{eq:eu} will become equivalent to  
\ifdefined\singlecolumn
\begin{align}
\min_{y'\in \cY'_I}
\EE_{p_M(\theta\mid z_{1:n})} \EE_{p(y\mid \theta,x=x_*)} r(y', y; x_*)  - 
\EE_{p_M(\theta\mid z_{1:n})} \min_{y'\in \cY'_I}\EE_{p(y\mid \theta,x=x_*)} r(y', y; x_*). \label{eq:eu-transformed}\tag{\ref{eq:eu}'}
\end{align}
\else
\begin{align}
&\min_{y'\in \cY'_I}
\EE_{p_M(\theta\mid z_{1:n})} \EE_{p(y\mid \theta,x=x_*)} r(y', y; x_*)  - \nonumber
\\
&\hspace{2em}\EE_{p_M(\theta\mid z_{1:n})} \min_{y'\in \cY'_I}\EE_{p(y\mid \theta,x=x_*)} r(y', y; x_*). \label{eq:eu-transformed}\tag{\ref{eq:eu}'}
\end{align}
\fi
The equivalence %
can be proved with the same idea as \citet{fong_martingale_2021}; we provide a proof in App.~\ref{app:proof-eu} for completeness. 
\eqref{eq:eu-transformed} represents %
the amount of reducible risk should we have full knowledge about the distribution of $y\mid x=x_*$, %
which can be determined given either the latent $\theta$ or an infinite amount of missing data. %
As noted in \citet{xu_minimum_2020}, 
This is precisely the intuition behind Bayesian epistemic uncertainty, and we recover standard epistemic uncertainty measures (e.g., posterior variance for a regression mean, or mutual information) if we return to the standard choices of $r$ such as the square or log loss.

Within the Bayesian framework, \eqref{eq:eu}
with a finite choice of $N$ will provide a lower bound for \eqref{eq:eu-transformed}: 
by convexity the second term will be larger. The intuition is that knowledge of the samples $z_{n+1:N}$ does not always enable the full reduction of risk. 
We can view the lower bound as an indication of the ``true'' epistemic uncertainty \eqref{eq:eu-transformed}, and it can be tightened if we choose $\{x_{n+i}\}$ to be more similar to $x_*$.  
The bound will also be tight for finite $N$ 
if the uncertainty is ``fully aleatoric'' or ``fully epistemic'', i.e., if the risk %
at $x=x_*$ cannot be reduced by any amount of additional $z_{n+1:N}$ 
or if it can be fully reduced by a single $z_{n+1}$. 
Alternatively, we can observe that we do not always need the tightest bound for \eqref{eq:eu-transformed}: if all we can do is to collect $N-n$ real samples, the remaining uncertainty %
will be effectively irreducible.

As discussed in \S 1, this Bayesian view may be justified through a common belief that high-capacity LMs may approximate Bayesian inference. 
It is also supported by the results of \citet[\S 2.6, \S 4]{wen_predictions_2022}, which imply that in a multi-task setting, any $p_M$ with a strong average-case performance for prediction %
will yield a similar value for \eqref{eq:eu}. %
\citet{falck2024are} investigated this belief empirically on synthetic numerical datasets and found it to be valid for smaller choices of %
$N$. %
Taken together, 
these results suggest that the Bayesian perspective can be relevant when we restrict $N$ to a regime where ICL is statistically efficient.
 
\vshrink{-0.3em}
\paragraph{Connection to non-Bayesian methods.}
Independent to the above, %
Eq.~\eqref{eq:eu} can also be understood through its connection to recent works. %
\citet{johnson_experts_2024,ahdritz_distinguishing_2024} proposed methods that 
measure the correlation between consecutive predictions $(y_{n+1}, y_*)$ for the same input $x_{n+1}=x_*$; 
as argued in \citet{johnson_experts_2024}, any learning system %
$p_M$ with complete confidence %
should lead to conditionally i.i.d.~$(y_*, y_{n+1})$ given $I$, and any $p_M$ with a ``fully epistemic'' uncertainty should produce $y_*=y_{n+1}$. It is clear that \eqref{eq:eu} matches these behaviours at the extremes: it equals 0 in the former case, %
and in the latter case correctly indicate a fully reducible risk. %
The Bayesian perspective provides additional insights to the works of \citet{johnson_experts_2024,ahdritz_distinguishing_2024}. 
Eq.~\eqref{eq:eu} generalizes their methodology by allowing for a wider range of choices for $\{x_{n+1:N}\}$, and the above discussion provides guidance on their choices.  

\begin{remark}[generalizations]\label{rem:eu-general}
\eqref{eq:eu} can be readily applied beyond ICL if we use $(x_{n+1:N}, y_{n+1:N})$ to represent more general queries, e.g., any clarifying questions and their answers. 
In such scenarios, \eqref{eq:eu} remains a non-negative measure of reducible uncertainty. 
While we will focus on ICL in experiments, we note that this generalization can be useful when the real responses are more expensive to obtain compared to LM generations, for example when they require interaction with user \citep{li_eliciting_2023} or a third-party source \citep{min2024silo}. 
\end{remark}

\section{Related Work}\label{sec:related-work}

\paragraph{Probabilistic UQ for LMs. } %
UQ has long been studied in %
LM applications involving a short or fixed-form response \citep[e.g.,][]{kamath-etal-2020-selective,jiang2021can}. 
Free-form generation introduces additional challenges. 
Our discussions %
are motivated by recent works on semantic uncertainty \citep{kuhn_semantic_2023,tian_just_2023}, which we seek to clarify and extend to a general scope.  
The measure \eqref{eq:mar} is %
a generalized entropy \citep{degroot1962uncertainty}; similar to \citet[p.8]{xu2020theory}, we deviate from that line of classical work by considering general loss functions and computational constraints. 

All methods reviewed so far can be viewed as adopting a \emph{probabilistic} perspective: they quantify uncertainty through the variation in the LM's predictive distribution. 
Another line of work proposes to prompt the LM for its %
uncertainty, which we discuss shortly. 
Orthogonal to these developments, conformal prediction methods \citep[e.g.,][]{quach_conformal_2023,mohri_language_2024,yadkori2024mitigating} %
take a prespecified uncertainty measure and a calibration set as input and provide a predictor with %
coverage guarantees.

\paragraph{Calibration in free-form generation. } \citet{band_linguistic_2024} studied calibration in scenarios where the free-form generation $y$ is used for downstream classification based on a known predictor $\phi(y)$. %
Our work generalizes their setup, as we can always define a utility for $y$ based on $\phi$ and the standard utility functions for classification. 
\citet{huang_uncertainty_2024} studied the \emph{rank-}calibration of general uncertainty measures which, as discussed in their work, %
is a different property. We note that an LM could be arbitrarily overconfident or underconfident for the uncertainty measure \eqref{eq:mar} to remain rank-calibrated. 
\citet{huang_calibrating_2024} proposed several calibration metrics based on a similarity measure. In their work both the similarity measure and the method to aggregate calibration error are left underspecified, rendering the resulted metrics hard to interpret; in contrast, our Eq.~\eqref{eq:gece} admits a straightforward interpretation as a generalized ECE. 

\paragraph{Prompt-based methods, instruction-tuned LMs.} 
A recent line of work proposes to quantify uncertainty through prompting, e.g., by asking the LM to output the probability that its generation is correct \citep{xiao2021hallucination,mielke2022reducing,pmlr-v239-ren23a} or to answer a multiple-choice question \citep{kadavath_language_2022}. Compared with the probabilistic methods, %
this approach can be more efficient computationally as it avoids the need to draw multiple samples for each query. 
However, it requires the LM to have stronger reasoning capabilities, and is more difficult to apply when the user's utility involves factors beyond factual correctness or for the quantification of epistemic uncertainty.

Still, within the above scope, it is reasonable to ask whether the verbalized approach might be preferable for instruction-tuned LMs. 
The question remains largely open: 
previous studies have found that instruction tuning can affect probabilistic calibration \citep{achiam2023gpt,tian_just_2023}, 
but empirical comparisons between probabilistic and verbalized approaches yielded inconsistent results \citep{kuhn_semantic_2023,tian_just_2023,xiong_can_2024}. 
Moreover, in the free-form setting there is also a lack of principled evaluation of calibration: 
many prior works \citep{kuhn_semantic_2023,lin2024generating} adopted evaluation metrics that conflate calibration with predictive performance \citep[p.5]{huang_uncertainty_2024}, 
and some studies \citep{jiang2021can,tian_just_2023} evaluated token-level (rather than task-specific) calibration, which is less relevant. 
Our experiments in \S\ref{sec:exp-qa} will fill in this gap.

\paragraph{Epistemic uncertainty.} 
Our method is inspired by \citet{fong_martingale_2021,xu_minimum_2020}. 
Similar ideas were discussed by \citet{berti_class_2021,wen_predictions_2022} in relation to \citet{fong_martingale_2021}, 
and \citet{grunwald_game_2004,kotelevskii_predictive_2024} in relation to \citet{xu_minimum_2020}. 
Similar notions of reducible uncertainty also appear in 
active learning \citep{smith2023prediction}, and in 
experiment design \citep{rainforth2024modern} in relation to \Cref{rem:eu-general}. 
All of these works either restricted to the estimation of a full %
data distribution, or required explicitly Bayesian models,\footnote{
i.e., a $p_M$ that corresponds to a Bayesian posterior predictive, 
defined by a known posterior over model parameters and a likelihood. 
}%
rendering them inapplicable to general LM applications. 
\citet{falck2024are} applied the method of \citet{fong_martingale_2021} to ICL, but similarly restricted to density estimation for numerical data and %
approximately Bayesian models. 
Concurrent work of \citet{yadkori2024believe} proposed a measure similar to \eqref{eq:eu}, but based on the log loss. It thus cannot distinguish between task-specific and irrelevant uncertainties, and also requires access to the LM's predictive density; our work avoids such limitations.

\section{Experiments}\label{sec:exp}

\subsection{Uncertainty in Question Answering}\label{sec:exp-qa}

\newcommand{\coqa}{\texttt{CoQA}\xspace}
\newcommand{\nqopen}{\texttt{NQOpen}\xspace}
\newcommand{\sciq}{\texttt{SciQ}\xspace}
\newcommand{\trivia}{\texttt{TriviaQA}\xspace}
\newcommand{\truthfulqa}{\texttt{TruthfulQA}\xspace}
\newcommand{\gptom}{\texttt{gpt-4o-mini}\xspace}
\newcommand{\gpt}{\texttt{gpt-4o}\xspace}
\newcommand{\llamaSmall}{\texttt{llama-3.1-8b}\xspace}
\newcommand{\llamaMid}{\texttt{llama-3.1-70b}\xspace}
\newcommand{\geminiPro}{\texttt{gemini-1.5-pro}\xspace}
\newcommand{\geminiFlash}{\texttt{gemini-1.5-flash}\xspace}
\newcommand{\claude}{\texttt{claude-3.5-sonnet}\xspace}

We first illustrate the uncertainty and calibration measures by revisiting the question answering (QA) experiments in recent works \citep{lin2024generating,tian_just_2023}. 
As discussed in \S\ref{sec:related-work}, 
our goal is to determine whether probabilistic UQ methods are suitable for modern instruction-tuned LMs, by providing a principled evaluation of task-specific calibration. %

\vshrink{-0.3em}
\paragraph{Experiment setup.} We combine the QA datasets studied by \citet{lin2024generating,tian_just_2023}: 
\coqa \citep{reddy2019coqa}, \trivia \citep{joshi2017triviaqa}, \nqopen \citep{nqopen}, \sciq \citep{welbl2017crowdsourcing}, and \truthfulqa \citep{lin2021truthfulqa}. 
We subsample 1000 queries for each dataset and report bootstrap confidence intervals (CIs) for all metrics. 
The utility $S(y',y;I)$ is defined by instructing an LM to rate the consistency of $y'$ with $y$ as an answer to the user's question. 
We adopt the prompt templates in \citet{lin2024generating} for evaluation and generation. 
We add a system prompt that instruct LMs to generate single-line responses, and use \gptom %
for evaluation.\footnote{
We confirmed that evaluation remains aligned with human perception, and that except on \sciq the generations are still varied in form. \sciq is a short-form dataset which we include for consistency with \citet{tian_just_2023}.
} 
As \trivia and \nqopen contain trivia questions that were annotated well before the knowledge cutoff date of current LMs, we further remove a subset of questions with clearly outdated answers; see App.~\ref{app:exp-qa-details} for details. 

We use LMs from the following model families: \gpt \citep{gpt4o}, 
\texttt{llama-3.1} \citep{dubey2024llama}, 
\texttt{gemini-1.5} \citep{geminiteam2024gemini}, and \texttt{claude-3.5} \citep{claude35}. 
All LMs are instruction tuned unless otherwise noted. 
In a subset of experiments, we compare 
the uncertainty measure \eqref{eq:mar} (denoted as \texttt{Prob.}) with the prompt-based methods of \citet[denoted as \texttt{P(True)}]{kadavath_language_2022} and %
\citet[\texttt{Verb.}]{tian_just_2023}. We apply 
all methods to the Gibbs predictor, and estimate uncertainty measures using Monte-Carlo estimation with 10 samples for each query. 
To evaluate calibration, we use the ECE metric \eqref{eq:gece} estimated through histogram binning \citep{naeini2015obtaining}. For reference, we also compute the average test utility to quantify predictive performance, as well as the AURAC metric from \citet{nadeem2009accuracy,lin2024generating}, which measures the performance of a selective prediction procedure based on a given LM and uncertainty measure. 
Full setup details are deferred to App.~\ref{app:exp-qa-details}. 

\vshrink{-0.3em}
\paragraph{Results and discussion.}
We defer full results to App.~\ref{app:exp-qa-results} and summarize the main findings below. 

\begin{figure}[htbp]
    \centering 
\ifdefined\singlecolumn
    \includegraphics[width=0.6\linewidth,clip,trim={0.2cm 0.57cm 0.2cm 0.47cm}]{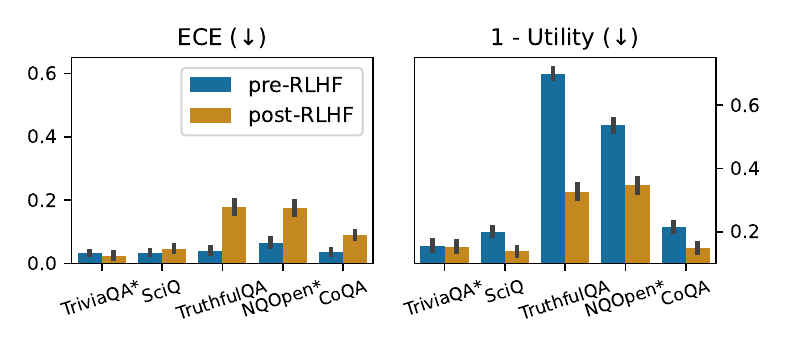}
\else
    \includegraphics[width=0.95\linewidth,clip,trim={0.2cm 0.57cm 0.2cm 0.47cm}]{figs/uqnlg/revisit-tian.pdf}
\fi
    \caption{Question answering: 
    ECE and utility for the \llamaMid model before and after instruction tuning. Error bar denotes 95\% bootstrap CI. 
    }\label{fig:revisit-tian}
\end{figure}

\paragraph{How does instruction tuning affect calibration? }
We revisit an experiment in \citet{tian_just_2023} on %
the impact of instruction tuning on \emph{token-level} calibration. 
Fig.~\ref{fig:revisit-tian} plots the ECE and test utility of \llamaMid before and after instruction tuning. Its left subplot can be compared with \citet[Fig.~1 left, for the first three datasets]{tian_just_2023}. 
The comparison reveals a rather different phenomenon: 
whereas the calibration of token-level probability deteriorated significantly (an increase in ECE by $0.1$--$0.2$) across all three datasets in \citet{tian_just_2023}, we find 
the deterioration of \emph{task-specific} calibration to be much smaller, except on \truthfulqa.
It is interesting to note that among these datasets, \truthfulqa is the only one where instruction-tuning led to a significant improvement in utility; additional results on \nqopen and \coqa further confirm this trend. 
These results indicate that \emph{the impact of instruction tuning on calibration may have been overestimated} in previous work. %
Results for AUARC are provided in App.~\ref{app:exp-qa-results}, where we also evaluate additional LMs. 
We find that across all datasets, the performance of selective prediction based on task-specific uncertainty is either 
approximately unchanged or improved after instruction tuning. 
This is similarly in contrast to \citet[Fig.~1 right]{tian_just_2023}, which shows that selective prediction based on token-level uncertainty consistently performs worse after instruction tuning.

\begin{figure}[htbp]
    \centering 
\ifdefined\singlecolumn
    \includegraphics[width=0.5\linewidth,clip,trim={0 0.47cm 0 0.57cm}]{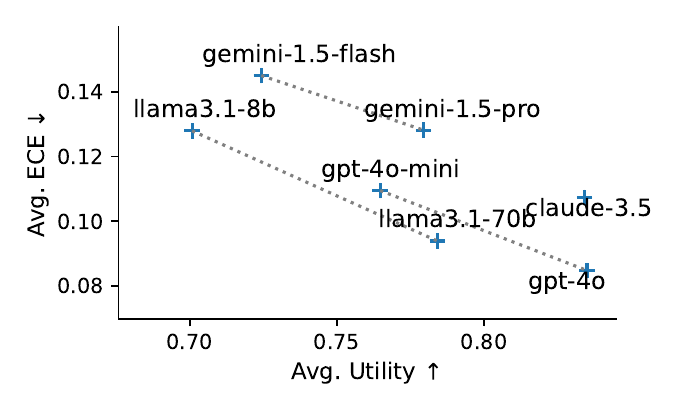}
\else
    \includegraphics[width=0.8\linewidth,clip,trim={0 0.47cm 0 0.57cm}]{figs/uqnlg/avg-ece-vs-utl.pdf}
\fi
    \caption{Question answering: test utility vs ECE for isntruction-tuned LMs, averaged over the datasets in \citet{lin2024generating}. 
    App.~\ref{app:exp-qa-results} presents results for individual datasets.
    }\label{fig:avg-ece-vs-utl}
\end{figure}

We now compare the calibration of instruction-tuned LMs with their predictive performance. 
As shown in Fig.~\ref{fig:avg-ece-vs-utl}, calibration is generally correlated with predictive performance, but substantial variation remains. %
It is thus reasonable to expect that the calibration of many instruction-tuned LMs may still be improved with different tuning methodologies.

\begin{figure*}
    \centering 
    \includegraphics[clip,trim={0.32cm 0.62cm 0.32cm 0.47cm},width=0.955\linewidth]{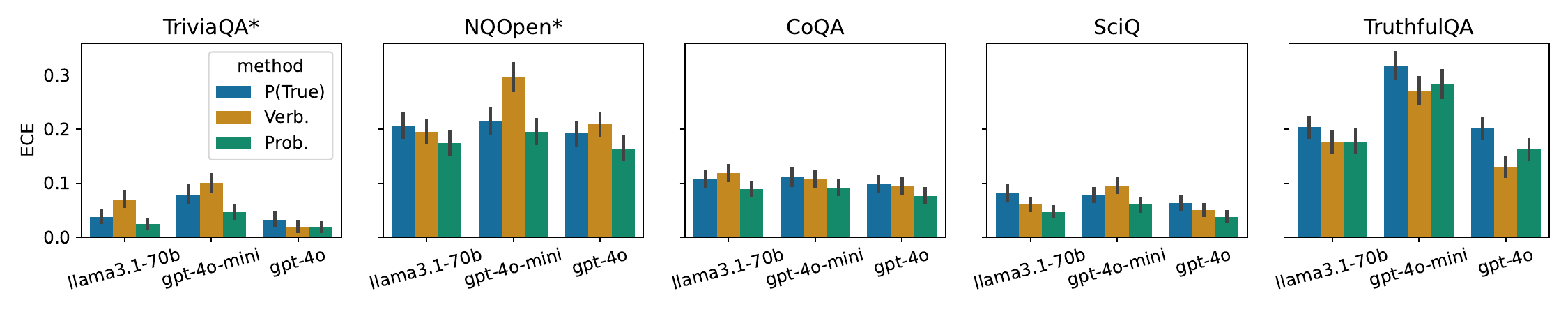}
    \caption{Question answering: calibration error from different methods. 
    Error bar denotes 95\% bootstrap CI.
    }\label{fig:qa-method-comparison}
\end{figure*}

\paragraph{Comparing probabilistic and prompt-based uncertainty. }
We compare the calibration of the uncertainty measure \eqref{eq:mar} with prompt-based alternatives, for which we can define an analogous ECE metric. %
Fig.~\ref{fig:qa-method-comparison} visualizes ECE from different methods. 
We can see that the measure \eqref{eq:mar} generally leads to the best calibration error; the only exception is on \truthfulqa, which is ``adversarially constructed'' to evaluate LMs' tendency to repeat human falsehoods. 
Reliability diagrams in App.~\ref{app:exp-qa-results} provide further insights. %
Taken together, the results suggest that \emph{the probabilistic approach to UQ remains effective for modern instruction-tuned LMs.}

Note that %
in the above comparison, 
only the probabilistic uncertainty measure makes use of the utility $S$. However, 
as discussed before, in many scenarios it is reasonable to assume access to (a surrogate of) $S$. Moreover, the ability of the probabilistic approach to incorporate different choices of utility can also be viewed as an advantage, especially when %
utility cannot be interpreted as a probability of correctness. 

\begin{figure*}[h]
    \centering 
    \includegraphics[clip,trim={0 0.49cm 0 0.17cm},width=0.97\linewidth]{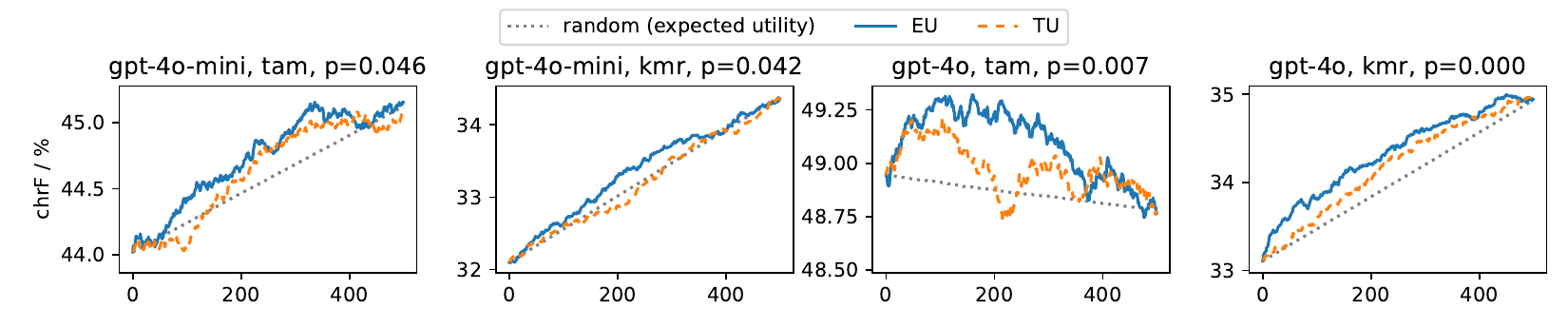}
    \caption{
Machine translation: average utility vs the number of deferrals to the many-shot predictor. 
We also report the p-value of a permutation test that compares the %
AUC-DF %
from the \texttt{EU} method to \texttt{random}. 
    }\label{fig:mt-main}
\vshrink{-0.15em}
\end{figure*}

\vshrink{-0.2em}
\subsection{Epistemic Uncertainty in ICL}\label{sec:exp-mt}
\vshrink{-0.2em}

We now turn to epistemic uncertainty (EU) quantification in ICL, and %
investigate whether the proposed EU measure may provide an effective deferral strategy that routes selected queries to a more expensive predictor. %

\vshrink{-0.3em}
\paragraph{Background and setup.} LM generations can often be improved given information retrieved from an external source. Such a %
process can become expensive if retrieval is from a third party \citep{min2024silo} or if it leads to a very long prompt \citep{agarwal_many-shot_2024}. %
In such scenarios, it would be desirable to limit retrieval to %
queries with a higher level of reducible uncertainty.
 
\citet{agarwal_many-shot_2024} showed that many-shot ICL can be effective %
in low-resource machine translation \citep{haddow2022survey}. Informed by their findings, we evaluate the use of EU for choosing between a LM predictor given few-shot demonstrations and a more expensive predictor based on many-shot ICL. We compare this method (denoted as \texttt{EU}) to a \texttt{random} baseline, and an alternative based on total uncertainty (\texttt{TU}, Eq.~\eqref{eq:mar}). 

We follow the setup in \citet{agarwal_many-shot_2024} and adopt the chrF score \citep{popovic2015chrf} as the utility. 
chrF is a lexical similarity score that measure both semantic and syntactic differences; %
the latter provides a source of irreducible uncertainty. 
We define a base predictor using ICL with %
$n=4$ shots, and a more expensive predictor with $n'=128$.  
We defer to the expensive predictor on $m$ test queries with the highest EU, and report the average utility (over all test samples) w.r.t.~$m$. Following \citet{gupta2024language}, we refer to this function as a \emph{deferral curve}, and use its integral %
(AUC-DF) as a scalar summary. 

The EU measure \eqref{eq:eu} is instantiated with $N-n=4$ and 
estimated as follows: 
for a query $I=\<z_{1:n}, x_*\>$, we first draw $l$ i.i.d.~samples $
\{z^{(i)}_{n+1:N} = 
(x^{(i)}_{n+1:N}, y^{(i)}_{n+1:N})\}_{i=1}^l$ where 
$x^{(i)}_{n+1:N}\sim p_x(\cdot\vert I)$ and $y^{(i)}_{n+1:N}\sim \prod_{k=n+1}^N p_M(y_k\mid \<z_{1:n},z_{n+1:k-1}^{(i)},x_k^{(i)}\>)$. Then, for each $i$, we sample %
$\{y^{(i,j)}_*\}_{j=1}^m \sim p_M(y_*\mid\<z_{1:n},z_{n+1:N}^{(i)}, x_*\>)$. We use $l(m-1)$ samples to define a plug-in Monte-Carlo estimator, 
and the remaining samples, $\{y^{(i,m)}_*\}_{i=1}^l$, to define the action space %
in \eqref{eq:eu}. We define 
$p_x$ by using an LM to rewrite %
$x_*$, and choose $l=5,m=8$.

Following \citet{agarwal_many-shot_2024}, we adopt 
the translation tasks from English to Kurdish (\texttt{kmr}) and Tamil (\texttt{tam}) in the 
FLORES+ dataset \citep{nllb-22}. Due to resource limitations, we subsample 500 data points from each dataset. 
We consider the following LMs: \gptom, \gpt and \geminiPro. 
Full setup details are deferred to 
Appendix~\ref{app:exp-mt-setup}. 

\vshrink{-0.3em}
\paragraph{Result and discussion.} Fig.~\ref{fig:mt-main} plots the deferral curves from the GPT models. We can see that the \texttt{EU} method consistently outperforms the \texttt{random} baseline, and the improvement in AUC-DF is statistically significant. In contrast, the \texttt{TU} baseline is not always better than random, which is expected since a larger number of in-context demonstrations will not help if predictive uncertainty is irreducible. 
To further compare between the \texttt{EU} and \texttt{TU} methods, we compute the bootstrap distributions of their AUC-DF metrics, which are obtained by resampling the test queries without replacement. As shown in Table~\ref{tbl:diff}, for all but one experiment (\geminiPro on \texttt{kmr}) \texttt{EU} outperforms \texttt{TU} in more than 75\% of bootstrap simulations. 

\begin{table}[htb]
    \centering
    \caption{Machine translation: AUC-DF difference between \texttt{EU} and \texttt{TU} methods. We report the median, $25\%$ and $75\%$ percentiles across 1000 bootstrap simulations. 
    }\label{tbl:diff}
    \vshrink{0.5em}
\ifdefined\singlecolumn\small
\else
    \resizebox{.95\linewidth}{!}{%
\fi
    \begin{tabular}{cccc} \toprule
        & gemini-1.5-pro & gpt-4o-mini & gpt-4o \\ \midrule
        tam 
             & $ 0.05 $ {\scriptsize $[0.00, 0.09]$}
             & $ 0.09 $ {\scriptsize $[0.01, 0.16]$}
             & $ 0.12 $ {\scriptsize $[0.06, 0.19]$} \\
        kmr 
             & $ 0.02 $ {\scriptsize $[-0.02, 0.06]$}
             & $ 0.15 $ {\scriptsize $[0.10, 0.21]$}
             & $ 0.15 $ {\scriptsize $[0.08, 0.22]$} \\
        \bottomrule \end{tabular}%
\ifdefined\singlecolumn
\else
    }
\fi
    \vshrink{-0.5em}
\end{table}

In the \geminiPro experiments, both \texttt{EU} and \texttt{TU} demonstrate less improvement over the random baseline. This may be attributable to the fact that uncertainty from \geminiPro is less calibrated: the respective %
base predictor achieves an average ECE of 0.303 on the two datasets, compared with 0.104 using \gptom and 0.083 using \gpt. Nonetheless, \texttt{EU} still appears to be the more effective method, as indicated by Table~\ref{tbl:diff}. See App.~\ref{app:exp-mt-results} for full results.

\section{Conclusion}\label{sec:conclusion}

This work studies uncertainty quantification in free-form natural language generation. 
Through a %
decision-theoretic perspective we derived principled methods to quantify the model's task-specific subjective uncertainty and to evaluate its calibration. 
These methods are also connected to previous work, which provides additional justification for their adoption. 
While the discussions are not necessarily novel outside the context of language modelling, within this domain the decision-theoretic perspective appears new and addresses important conceptual challenges in uncertainty quantification. 
Experiments demonstrated the practical utility of the proposed methods.

\bibliography{main}
\bibliographystyle{apalike}

\newpage
\appendix
\onecolumn

\section{Deferred Proof}\label{app:proof-eu}

\begin{claim}
Suppose $\cX,\cY$ are finite sets, and $\sup_{y',y,x}|r(y',y; x)|<\infty$. %
Suppose 
there exists a Bayesian model with prior $\pi$ over a discrete parameter $\theta\in\Theta$ and a (conditional) likelihood function $p(y\mid x,\theta)$, s.t.~for all $(n, x_{1:n}, y_{1:n}, x_*)$ we have $$
p_M(y_{n+1}=\cdot \mid x_{n+1}=x_*, x_{1:n}, y_{1:n}) = 
\int \pi(d\theta\mid x_{1:n}, y_{1:n}) p(y=\cdot\mid x=x_*, \theta)
$$ where $\pi(d\theta\mid x_{1:n}, y_{1:n}) \propto \pi(d\theta) \prod_{i=1}^n p(y=y_i\mid x=x_i,\theta)$ denotes the parameter posterior. 
Suppose \eqref{eq:eu} is defined using $x_{n+1:N}\overset{i.i.d.}{\sim} p_{x,1}$, and that 
for all $x_*\in\cX$ and $\pi$-a.e.~$\theta$ 
we have 
\begin{equation}\label{eq:lh-consistency}
\lim_{n\to\infty} \EE_{x_{1:n}\sim p_{x,1}, y_i\sim p(y=\cdot\mid x=x_i, \theta)} 
    \|p_M(y_{n+1}=\cdot\mid x_{n+1}=x_*, x_{1:n}, y_{1:n}) - p(y=\cdot\mid x=x_*, \theta)\|_{\ell_2(\cY)}^2 = 0,
\end{equation}
where $\|f\|_{\ell_2(\cY)} := \sqrt{\sum_{y\in\cY} f(y)^2}$ denotes the $\ell_2$ norm. 
Then for $p_{x,1}$-a.e.~$(x_{1:n},x_*)$ and $p_M$-a.e.~$y_{1:n}$, 
the $n\to\infty$ limit of \eqref{eq:eu} is equivalent to \eqref{eq:eu-transformed}. 
\end{claim}
In the above, \eqref{eq:lh-consistency} is the assumed identifiability condition for the likelihood function. 
We expect the numerous technical restrictions to be relaxable, but refrain from a more general proof for brevity.
The main conditions that enable the equivalence are that \emph{(i)} $p_M$ can be augmented to define an exchangeable (or c.i.d.) model for $(x,y)$, and \emph{(ii)} the likelihood function is identifiable. 

\begin{proof}
We will show that each of the two terms in \eqref{eq:eu} and \eqref{eq:eu-transformed} are equivalent. 

Let us introduce the following notations: define $\cZ:=\cX\times \cY$ and, for any $x_i,y_i$, $z_i:=(x_i,y_i)\in\cZ$. 
For all $n\ge 0$, define 
$
\bar p_M(z_{n+1}=(x,y) \mid z_{1:n}) = p_{x,1}(x) p_M(y_{n+1}=y\mid x_{n+1}=x, x_{1:n}, y_{1:n}), 
\bar p(z=(x,y)\mid \theta) = p_{x,1}(x)p(y\mid \theta,x)$. Clearly, $\bar p_M$ is equivalent to %
a Bayesian model with prior $\pi$ and the factorized likelihood $\bar p$. Thus,  
$z_{n+1:N+1}\sim \bar p_M(\cdot\mid z_{1:n})$ are exchangeable, and we have, for all $n, z_{1:n}, N>n, y'\in\cY$ and $p_{x,1}$-almost every $x_*$,  
$$
\EE_{\bar p_M}(r(y', y_{N+1}; x_*)\mid z_{1:n}, x_{N+1}=x_*) = 
\EE_{\bar p_M}(r(y', y_{n+1}; x_*)\mid z_{1:n}, x_{n+1}=x_*).
$$
By definitions, the terms above equal the first term in \eqref{eq:eu} and \eqref{eq:eu-transformed}, respectively, with their outmost infimum removed. 
Retaking infimum %
proves the equivalence of the first terms. 

For the second term, %
observe that by %
the boundedness of $r$ and the discreteness of $\cY$, for all $x_*\in\cX$ the function $R_{x_*}(\rho) := \inf_{y'\in\actionSpace[I]}\EE_{y\sim\rho} r(y',y; x_*)$ is Lipschitz continuous w.r.t.~the $\ell_2(\cY)$ norm.\footnote{Note that any distribution $\rho$ over $\cY$ can be identified with a probability mass function in $\ell_2(\cY)$ since $\cY$ is discrete.} %
Thus, for $\pi$-a.e.~$\theta$ we have 
$$
\lim_{N\to\infty} \EE_{z_{1:N-n}\overset{i.i.d.}{\sim}\bar p(z\mid \theta)} 
|R_{x_*}(p_M(y=\cdot\mid x=x_*,z_{1:N-n})) - R_{x_*}(p(y=\cdot\mid x=x_*,\theta))| = 0
$$
following \eqref{eq:lh-consistency}. 
It follows by the boundedness of $R_{x_*}(\cdot)$ and the dominated convergence theorem that 
\begin{align*}
&\phantom{=} \lim_{N\to\infty} \EE_{\theta\sim \pi,z_{1:N-n}\sim\bar p(z\mid \theta)} R_{x_*}(p_M(y=\cdot\mid x=x_*,z_{1:N-n})) %
=  \EE_{\theta\sim \pi}R_{x_*}(p(y=\cdot\mid x=x_*,\theta)). \numberthis\label{eq:eu-t2}
\end{align*}
The RHS above equals the expectation of the second term in \eqref{eq:eu-transformed} over $z_{1:n}\sim\bar p_M$. 
Denote the second term in \eqref{eq:eu} as $u_N(z_{1:n})$. By definition of $\bar p_M$ and its exchangeability, as well as the convexity of $R_{x_*}(\cdot)$, we have 
\begin{align*}
\EE_{\bar p_M(z_{1:n})} u_N(z_{1:n}) &= 
    \EE_{\bar p_M(z_{1:n})}\EE_{\bar p_M(z_{n+1:N}\mid z_{1:n})}  R_{x_*}(\bar p_M(y_{N+1}=\cdot\mid x_{N+1}=x_*, z_{1:N})) \\ 
&\le 
\EE_{\bar p_M(z_{n+1:N})}  R_{x_*}(\EE_{\bar p_M(z_{1:n}\mid z_{n+1:N})}\bar p_M(y_{N+1}=\cdot\mid x_{N+1}=x_*, z_{1:N}))
\\
&= \EE_{\bar p_M(z_{1:N-n})}  R_{x_*}(\bar p_M(y_{N-n+1}=\cdot\mid x_{N-n+1}=x_*, z_{1:N-n})) \\
&= 
\EE_{\theta\sim\pi,z_{1:N-n}\sim\bar p(z\mid \theta)} R_{x_*}(p_M(y=\cdot\mid x=x_*,z_{1:N-n})).
\end{align*}
Combining with \eqref{eq:eu-t2} and applying the dominated convergence theorem yield 
\begin{align}\label{eq:eu-t3}
\EE_{\bar p_M(z_{1:n})}\lim_{N\to\infty} u_N(z_{1:n}) = 
\lim_{N\to\infty} \EE_{\bar p_M(z_{1:n})} u_N(z_{1:n}) 
\le 
\EE_{\bar p_M(z_{1:n})}
\EE_{\pi(\theta\mid z_{1:n})}R_{x_*}(p(y=\cdot\mid x=x_*,\theta)).
\end{align}
On the other hand, we have 
\begin{align*}
u_N(z_{1:n}) &= \EE_{\bar p_M(z_{n+1:N}\mid z_{1:n})} R_{x_*}(\bar p_M(y=\cdot\mid x=x_*, z_{1:N}))  
\\  &
= \EE_{\bar p_M(z_{n+1:N}\mid z_{1:n})} R_{x_*}(\EE_{\pi(\theta\mid z_{1:N})} p(y=\cdot\mid x=x_*, \theta)) \\ 
&\ge \EE_{\bar p_M(z_{n+1:N}\mid z_{1:n})} \EE_{\pi(\theta\mid z_{1:N})}R_{x_*}( p(y=\cdot\mid x=x_*, \theta)) \\ 
&= \EE_{\pi(\theta\mid z_{1:n})}R_{x_*}( p(y=\cdot\mid x=x_*, \theta)),
\end{align*}
so the same holds for $\lim_{N\to\infty} u_N(z_{1:n})$. Comparing with \eqref{eq:eu-t3} we find that equality must hold for $\bar p_M$-a.e.~$z_{1:n}$. This proves the equivalence for the second terms in \eqref{eq:eu} and \eqref{eq:eu-transformed}, and consequently the original claim. 
\end{proof}

\section{Experimental Details}\label{app:exp-details}

Code and model samples for all experiments are available at \url{https://github.com/meta-inf/suq-nlg}.

\subsection{Question Answering: Deferred Setup Details}\label{app:exp-qa-details}

We adopt the prompts in \citet{lin2024generating} for generation and evaluation; see their Appendix B. For evaluation, their prompt instructs the LM to ``rate the level of consistency'' between a provided answer and a reference based on in-context demonstrations. 
For generation, we add a system prompt that instructs the LM to generate single-line responses to avoid interference with the evaluation prompt template in \citet{lin2024generating}; we find that without the system prompt, the more recent LMs we have evaluated have a greater tendency to generate multi-line responses. 
The generations can still contain multiple sentence, and we find them to be varied in format and content. 

The \nqopen and \trivia datasets contain a number of outdated questions, for example about the current holder of a public office or the latest season of an ongoing TV series. We thus use an LM (\gpt) to identify and remove such questions. We provide the LM with a question and its (possibly outdated) reference answer, and asks it to evaluate whether the question may fall into the above categories; the LM is asked to deprioritize its internal knowledge over the reference answer during evaluation. 
The prompt is shown in Fig.~\ref{fig:prompt-qa-filtering}. 
We note that \emph{the filtering process does not affect any finding reported in the main text}; see e.g.~Fig.~\ref{fig:revisit-tian-full} (a) and (b).
we chose to apply this process so that the reported results more accurately reflect the LMs' performance on these tasks.

We access all LMs through API services, from OpenAI (for GPT models) or Google Cloud. 
The API versions are as follows: 
\texttt{gpt-4o-2024-08-06}, \texttt{gpt-4o-mini-2024-07-18}, 
\texttt{gemini-1.5-flash-001}, \texttt{gemini-1.5-pro-001}, 
\texttt{claude-3.5-sonnet@20240620} (versioning from Google cloud). 
We use the default sampling hyperparameters which always include a temperature of 1. 

We implement the prompt-based uncertainty measures as follows: for the method of \citet{kadavath_language_2022} we adopt their prompt template and include model generations in the prompt, as recommended in their work; 
for the method of \citet{tian_just_2023}, we use the two-stage variant (\texttt{Verb.~2S top-1}) which separates answer and confidence generation, so that the answer generation process can be comparable with the other methods; we use their prompt for confidence generation. 
For all test metrics we report bootstrap CIs computed using 10000 simulations. 

All prompts and an illustrative subset of model generations can be found at the code repository.

\begin{figure}[htbp]
\begin{lstlisting}
    I will provide you with a trivia question and its reference answer as of {year}. Please evaluate if the answer to this question is likely to have changed between {year} - 2023. 

    Examples of questions that might have changed answers include:
    * Recent events or records
    * The latest season of an ongoing TV series
    * Current holders of public offices or titles
    
    Use your best judgment in the evaluation. In scenarios where your internal knowledge could be inaccurate, rely on common sense and the provided reference answer. Please format your response in two lines, as follows:
    1. A brief explanation of your reasoning (one sentence)
    2. A "Yes" or "No" answer indicating whether the answer likely changed
    
    Please adhere to the format of the example below:
    
    Question: when did the botswana currency first come into circulation 
    Reference Answer: 1976
    Reasoning: The date a currency first entered circulation is a historical fact that doesn't change.
    Answer: No
    
    Now please answer the following:
    
    Question: {question}
    Reference Answer: {answer}
\end{lstlisting}
\caption{Question answering: prompt template used to filter trivia datasets.}\label{fig:prompt-qa-filtering}
\end{figure}

\subsection{Question Answering: Additional Results}\label{app:exp-qa-results}

Full results in the setting of Fig.~\ref{fig:revisit-tian}--\ref{fig:qa-method-comparison} are provided in Fig.~\ref{fig:revisit-tian-full}--\ref{fig:qa-mc-full}, respectively, where we also provide results for AUARC. As expected, across LMs with a similar level of predictive performance, AUARC is generally correlated with calibration error: see the comparison between \llamaMid and \geminiPro (or \gptom) in Fig.~\ref{fig:aevu-full}. 
It is also interesting to note that in the setting of Fig.~\ref{fig:revisit-tian}, the calibration of the \texttt{gemma2-27b} model \citep{team2024gemma} often improves after instruction tuning; 
inspection on \trivia and \sciq appears to show that 
the instruction-tuned LM has a similar level of variability in its generations, but the quality of generation has generally improved. 

Fig.~\ref{fig:qa-rd} compares the reliability diagrams obtained from different uncertainty measures, evaluated on \llamaMid and \gpt. 
We can see that all uncertainty measures can be overconfident on inputs with low uncertainty, while the prompt-based measures can also be underconfident on high-uncertainty inputs.

\begin{figure}[htbp]
\centering
\subfigure[\llamaSmall]{\includegraphics[width=0.72\linewidth]{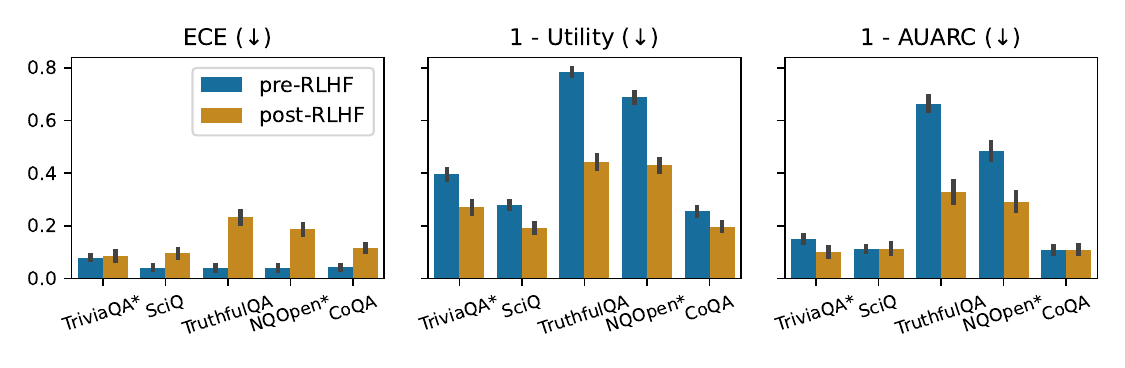}}
\subfigure[\texttt{gemma2-27b}]{\includegraphics[width=0.72\linewidth]{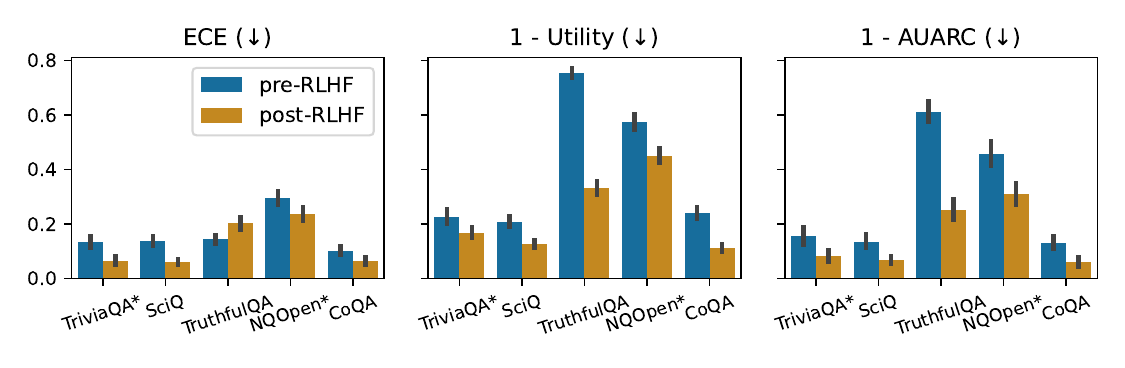}}
\subfigure[\llamaMid]{\includegraphics[width=0.72\linewidth]{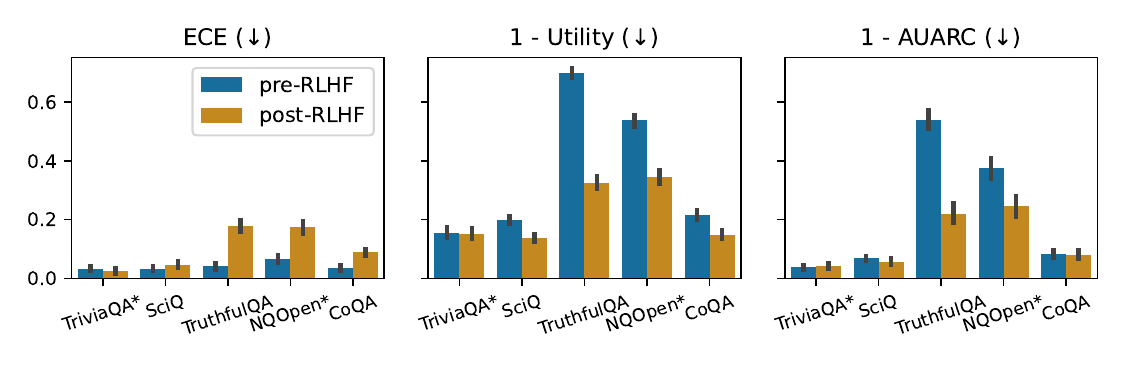}}
\subfigure[\llamaMid, without filtering for trivia datasets]{\includegraphics[width=0.72\linewidth]{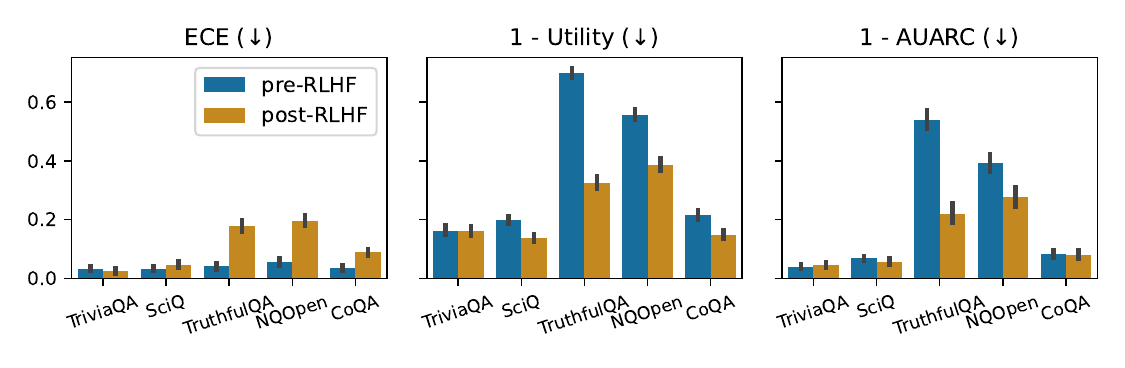}}
\caption{Question answering: full results in the setting of Fig.~\ref{fig:revisit-tian}.}\label{fig:revisit-tian-full}
\end{figure}

\begin{figure}[htbp]
\centering 
\includegraphics[width=0.95\linewidth]{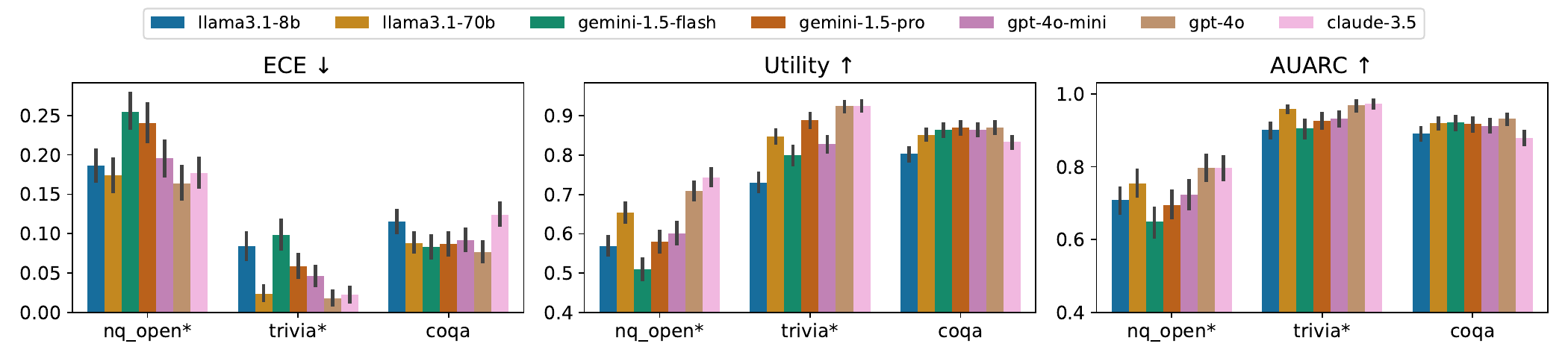}
\caption{Question answering: full results in the setting of Fig.~\ref{fig:avg-ece-vs-utl}. Error bar denotes 95\% bootstrap CI.}    \label{fig:aevu-full}
\end{figure}

\begin{figure}[htbp]
\centering 
\includegraphics[width=0.95\linewidth]{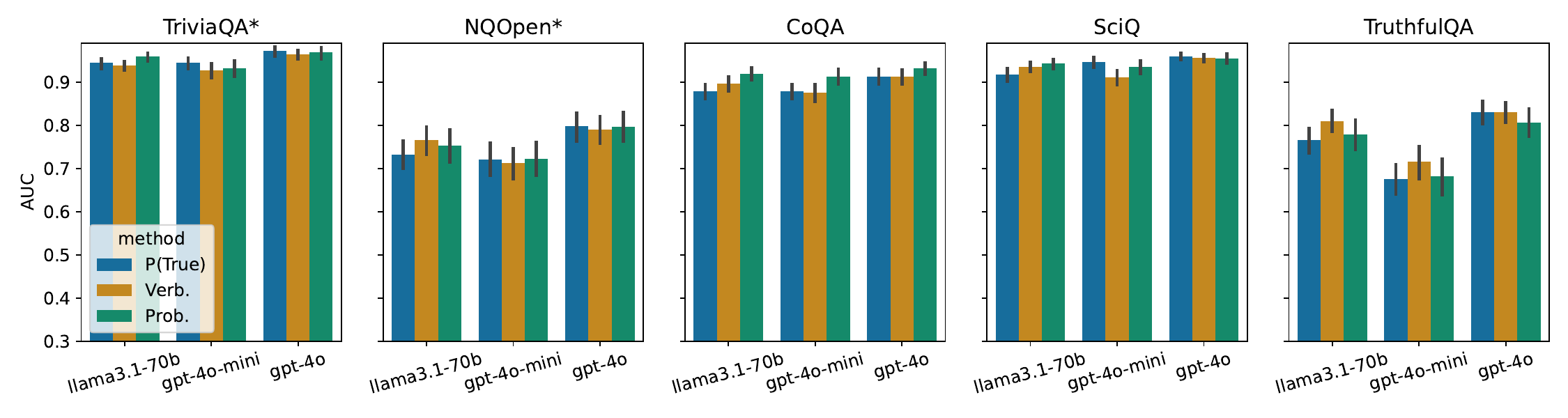}
\caption{Question answering: results for AUARC in the setting of Fig.~\ref{fig:qa-method-comparison}.}\label{fig:qa-mc-full}
\end{figure}

\begin{figure}[htbp]
\centering 
\subfigure[\llamaMid, \nqopen]{\includegraphics[width=0.47\linewidth]{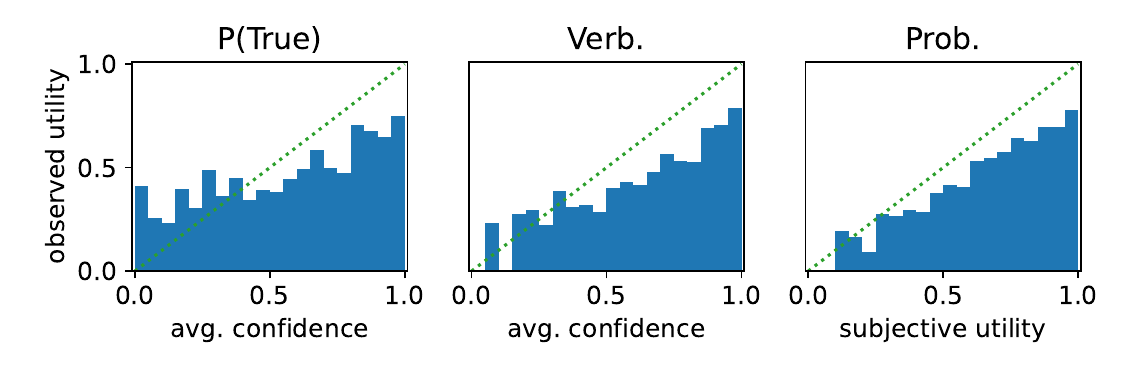}}
\subfigure[\gpt, \nqopen]{\includegraphics[width=0.47\linewidth]{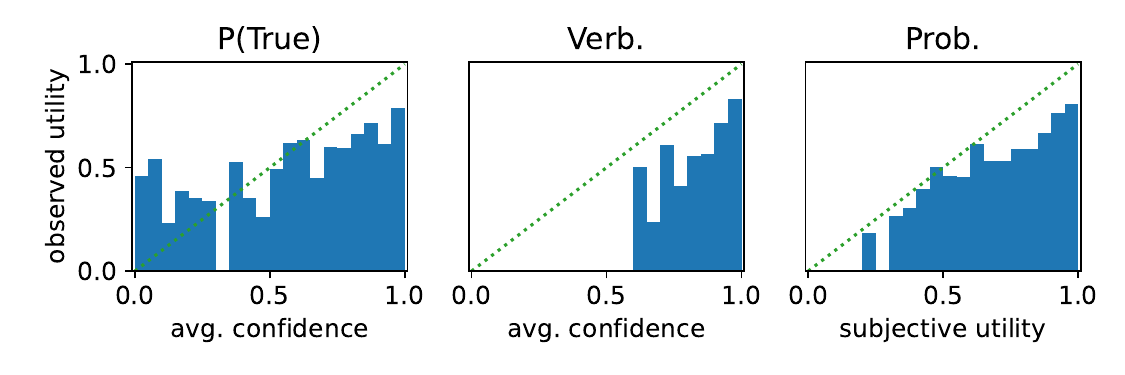}}
\subfigure[\llamaMid, \trivia]{\includegraphics[width=0.47\linewidth]{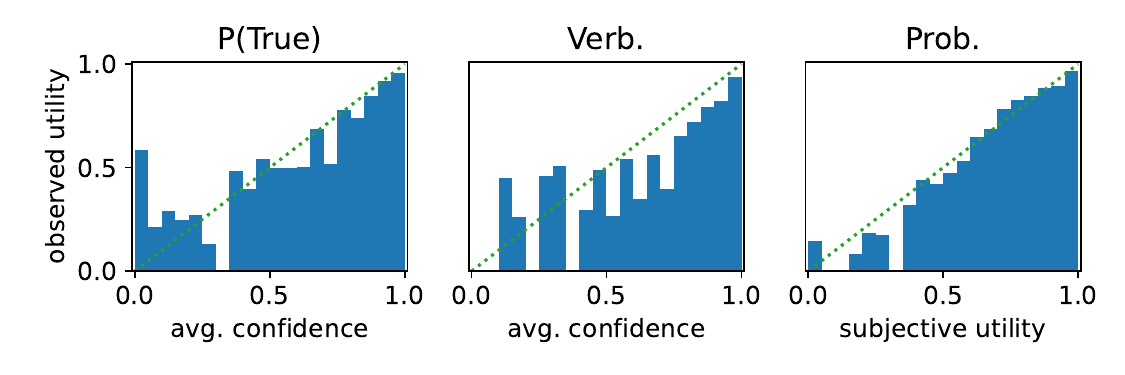}}
\subfigure[\gpt, \trivia]{\includegraphics[width=0.47\linewidth]{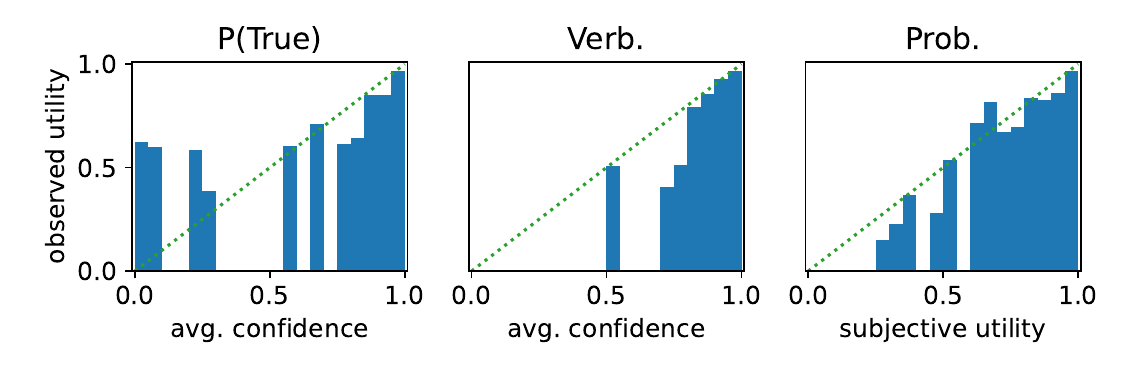}}
\subfigure[\llamaMid, \coqa]{\includegraphics[width=0.47\linewidth]{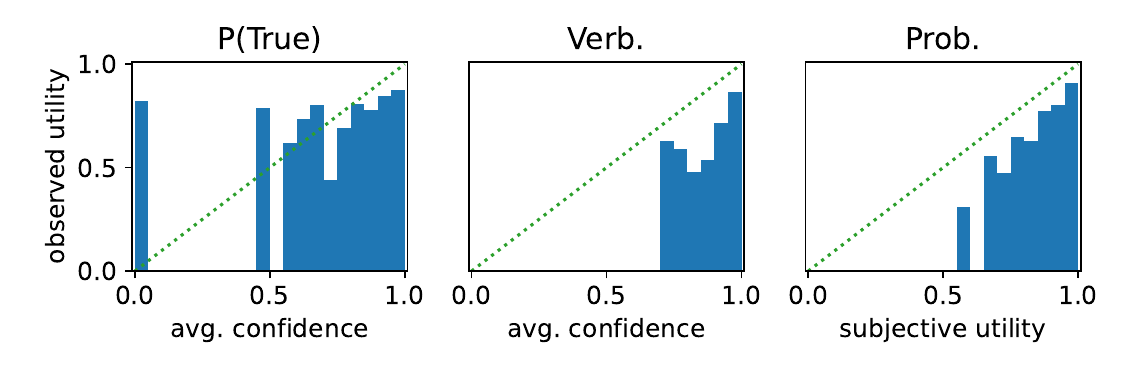}}
\subfigure[\gpt, \coqa]{\includegraphics[width=0.47\linewidth]{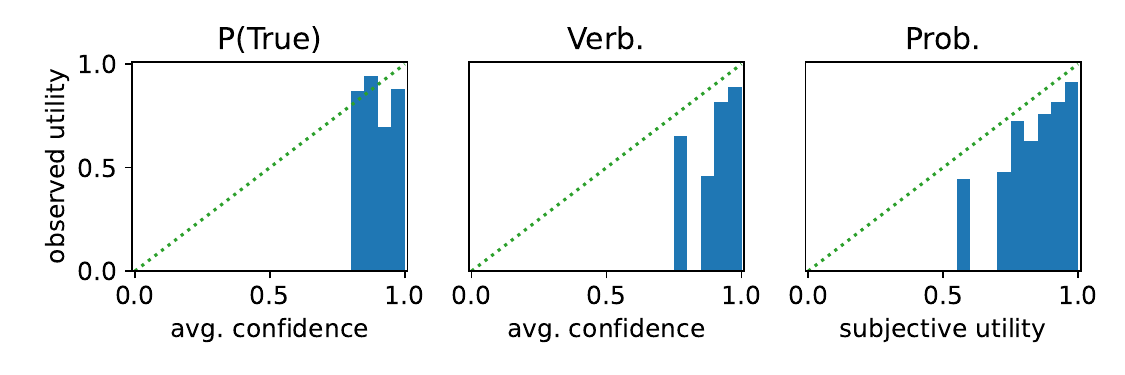}}
\subfigure[\llamaMid, \sciq]{\includegraphics[width=0.47\linewidth]{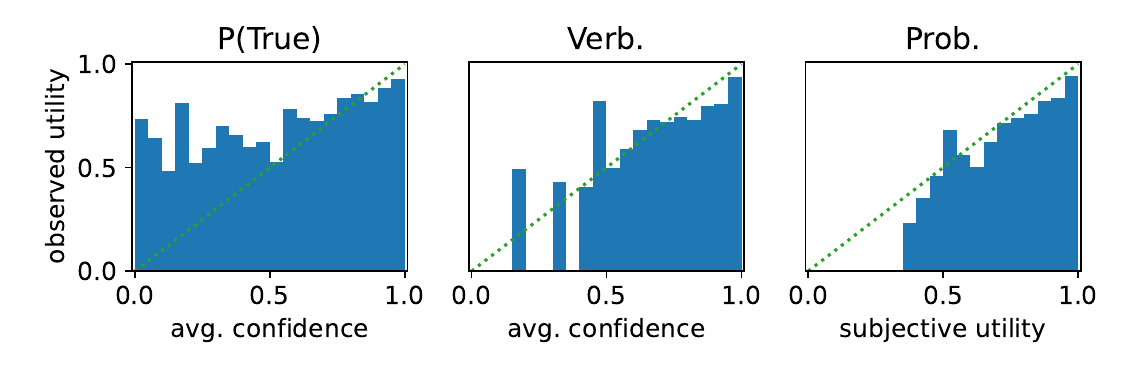}}
\subfigure[\gpt, \sciq]{\includegraphics[width=0.47\linewidth]{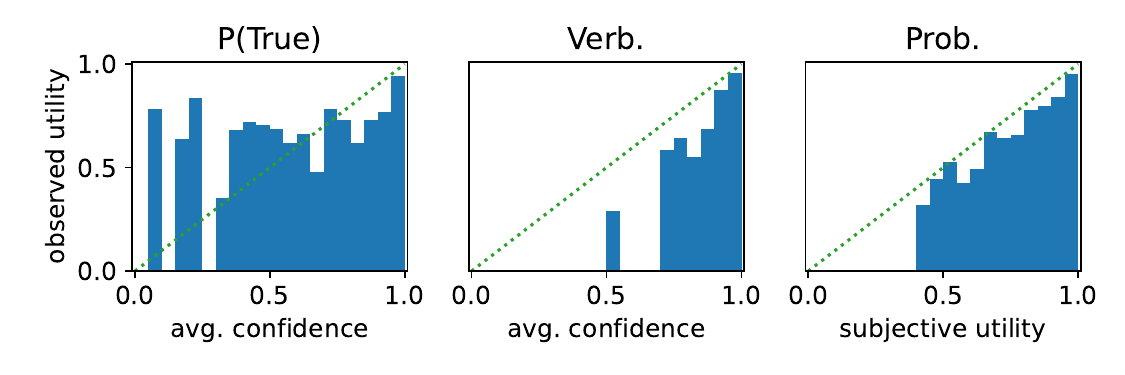}}
\subfigure[\llamaMid, \truthfulqa]{\includegraphics[width=0.47\linewidth]{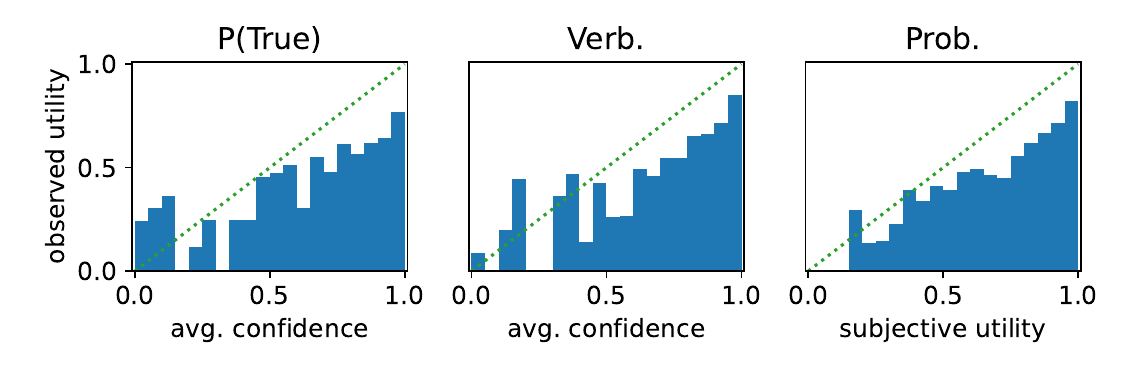}}
\subfigure[\gpt, \truthfulqa]{\includegraphics[width=0.47\linewidth]{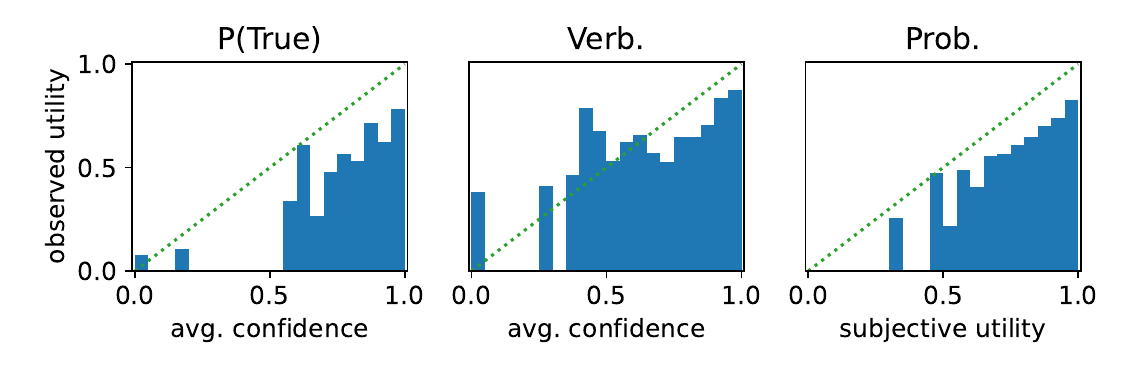}}
\caption{Question answering: reliability diagrams for the \gpt and \llamaMid models. 
Note that the $x$-axis is always equivalent to the negated uncertainty measure.
}\label{fig:qa-rd}    
\end{figure}

\subsection{Machine Translation: Setup Details}\label{app:exp-mt-setup}

We use the prompt in \citet{agarwal_many-shot_2024} for the translation task, and generate $x_{n+1:N}$ using \geminiFlash with the prompt in Fig.~\ref{fig:mt-prompt}. 
Details for the LMs involved are the same as App.~\ref{app:exp-qa-details}. 

The deferral experiments involve two ICL predictors. The base predictor is implemented by 
sampling $n=4$ demonstrations uniformly without replacement. For the more expensive predictor 
we append $n'-n=124$ demonstrations that are closest to the test query in an embedding space, 
where embeddings are computed using the \texttt{text-embedding-004} model from Google Cloud.\footnote{\url{https://cloud.google.com/vertex-ai/generative-ai/docs/model-reference/text-embeddings-api}} 
In both cases we retrieve demonstrations from the \texttt{dev} split of the dataset and evaluate 
on the \texttt{devtest} split.
We experimented with both (nontrivial) MBR and Gibbs decoding;\footnote{Note that these predictors do not coincide with the two predictors in \eqref{eq:eu}. Instead, \eqref{eq:eu} is only used as a surrogate to approximate the performance improvement in the deferral process.} 
the former is implemented by drawing $16$ samples for each query, using half of them to define $\actionSpace$ and half to approximate the expectation in \eqref{eq:expected-loss}. 
The main text reports the results for the MBR predictors, as we find them to perform better across all experiments. However, the same EU measure also works well for the Gibbs predictors, as shown in Fig.~\ref{fig:mt-gibbs}. 

Following \citet{robinson2023chatgpt}, we use the chrF2++ score implemented by \citet{post2018call}. 
The p-values reported in Fig.~\ref{fig:mt-main} are calculated using 10000 replications for the \texttt{random} baseline.

\begin{figure}[htbp]
\begin{lstlisting}
Please generate {K} sentences that convey a broadly similar meaning as the one provided, ensuring they are of similar length and use varied wording. Place a blank line after each sentence. Avoid adding any extra explanations.

{x_*}
\end{lstlisting}
\caption{Machine translation: prompt used to define $p_x$. $x_*$ denotes the test query.
}\label{fig:mt-prompt}
\end{figure}

\subsection{Machine Translation: Additional Results}\label{app:exp-mt-results}

\Cref{tbl:mt-ece} reports the calibration error from all models. We can see that \geminiPro consistently demonstrates worse calibration.

\begin{table}[htb]
    \centering
    \caption{Machine translation: ECE for all models.}\label{tbl:mt-ece}
\vspace{1ex}
    \begin{tabular}{cccc}
        \toprule 
        & \geminiPro & \gptom & \gpt \\ 
        \midrule 
        Tam & 0.250 & 0.041 & 0.065 \\
        Kmr & 0.356 & 0.017 & 0.010 \\ 
        \bottomrule
    \end{tabular}
\end{table}

\begin{figure}[htb]
\centering 
\subfigure[English $\to$ Tamil]{\includegraphics[width=0.24\linewidth]{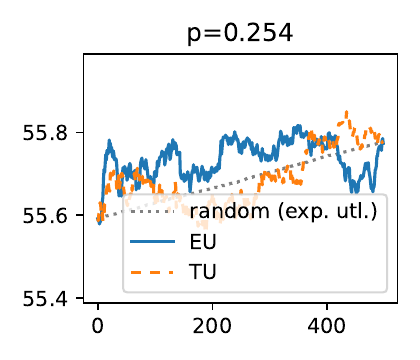}}
\subfigure[English $\to$ Kurdish]{\includegraphics[width=0.24\linewidth]{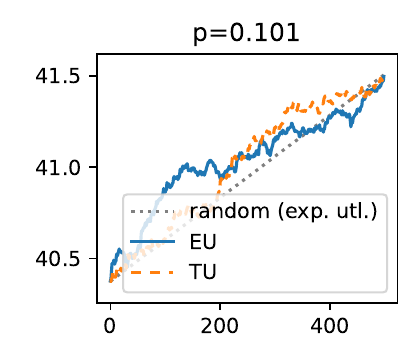}}
\caption{Machine translation: results for \geminiPro in the setting of Fig.~\ref{fig:mt-main}.}\label{fig:mt-full}
\end{figure}

\begin{figure}[htb]
\centering 
\subfigure[\geminiPro]{\includegraphics[width=0.24\linewidth]{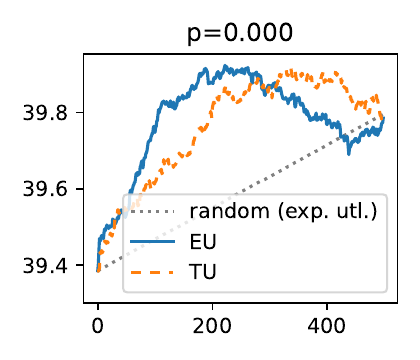}}
\subfigure[\gptom]{\includegraphics[width=0.24\linewidth]{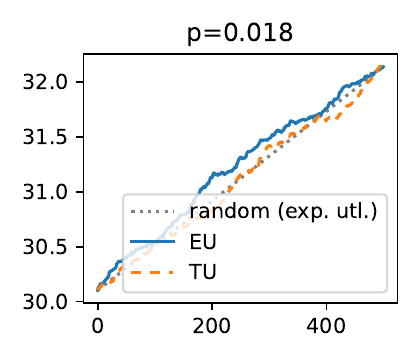}}
\subfigure[\gpt]{\includegraphics[width=0.24\linewidth]{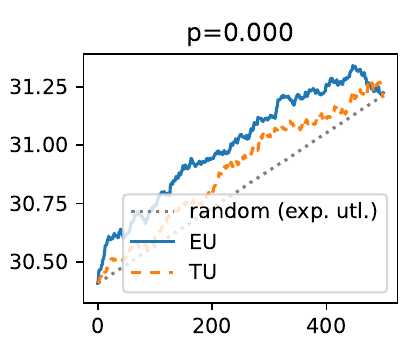}}
\caption{Machine translation: deferral curve for Gibbs predictors on the English $\to$ Kurdish task.}\label{fig:mt-gibbs}
\end{figure}

\end{document}